\documentclass[journal]{IEEEtran}

\ifCLASSINFOpdf
  
\else
 
\fi

\usepackage{graphicx}
\usepackage{cite}
\usepackage{float}
\usepackage{caption}
\usepackage{amsmath,amssymb,amsfonts}
\usepackage{textcomp}
\usepackage[export]{adjustbox}
\usepackage{bbm}
\usepackage{multirow, makecell}
\usepackage{array}
\usepackage{bbm}
\usepackage{booktabs, multirow} % for borders and merged ranges
\usepackage{soul}% for underlines
\usepackage{epstopdf}
\usepackage[section]{placeins}
\usepackage{enumitem}
\newlist{steps}{enumerate}{1}
\setlist[steps, 1]{label = \arabic*:}
\usepackage{mdwmath,mathrsfs}         
\usepackage{mdwtab}
\usepackage{paralist,fancyref,float}
\usepackage{amsthm}
\usepackage{tikz}
\usepackage{dsfont}
\usepackage{multirow}
\usepackage{mathtools, cuted}
\usepackage{algorithm}
\usepackage{algpseudocode}
\usepackage{tabularx}

\usepackage{hhline}
\usepackage{bbm}
\usepackage{amssymb}
\usepackage{times}
\usepackage{epsfig}
\usepackage{eso-pic}
\usepackage{xspace}
\usepackage{enumitem}
\usepackage{sidecap}
\usepackage{mathtools}
\usepackage{verbatim}
\usepackage{float}
\usepackage{stfloats}
\begin{document}

\title{Robust Semi-Supervised Learning for Histopathology Images through Self-Supervision Guided Out-of-Distribution Scoring}
% Self-Supervised Outlier Detection Framework for Open Set Semi-Supervised Learning in Digital Histology Images

\author{Nikhil~Cherian~Kurian,
        Varsha~S,~Abhijit~Patil,~Shashikant~Khade,
        and~Amit~Sethi% <-this % stops a space
\thanks{NC Kurian, S Varsha, A Patil, S Khade and A Sethi are with the Department
of Electrical Engineering, Indian Institute of Technology Bombay, Mumbai,
MH, 400076 India e-mail: nikhilkurian@iitb.ac.in.}% <-this % stops a space

% \thanks{Manuscript received April 19, 2005; revised August 26, 2015.}
}

\markboth{Arxiv Preprint}%
{Shell \MakeLowercase{\textit{et al.}}: Self-Supervision Guided Out of Distribution Score for Robust Semi-Supervised Learning in Histopathology Images}

\maketitle

\begin{abstract}
Semi-supervised learning (semi-SL) is a promising alternative to supervised learning for medical image analysis when obtaining good quality supervision for medical imaging is difficult. However, semi-SL assumes that the underlying distribution of unaudited data matches that of the few labeled samples, which is often violated in practical settings, particularly in medical images. The presence of out-of-distribution (OOD) samples in the unlabeled training pool of semi-SL is inevitable and can reduce the efficiency of the algorithm. Common preprocessing methods to filter out outlier samples may not be suitable for medical images that involve a wide range of anatomical structures and rare morphologies. In this paper, we propose a novel pipeline for addressing open-set supervised learning challenges in digital histology images. Our pipeline efficiently estimates an OOD score for each unlabelled data point based on self-supervised learning to calibrate the knowledge needed for a subsequent semi-SL framework. The outlier score derived from the OOD detector is used to modulate sample selection for the subsequent semi-SL stage, ensuring that samples conforming to the distribution of the few labeled samples are more frequently exposed to the subsequent semi-SL framework. Our framework is compatible with any semi-SL framework, and we base our experiments on the popular Mixmatch semi-SL framework. We conduct extensive studies on two digital pathology datasets, Kather colorectal histology dataset and a dataset derived from TCGA-BRCA whole slide images, and establish the effectiveness of our method by comparing with popular methods and frameworks in semi-SL algorithms through various experiments.
\end{abstract}

\begin{IEEEkeywords}
Semi Supervised learning, open-set, label-noise, mixmatch
\end{IEEEkeywords}

\IEEEpeerreviewmaketitle

\section{Introduction}
Medical image analysis requires large volumes of supervised data to train deep learning models effectively, but obtaining good quality supervision for medical imaging is inherently difficult due to the associated labor, expertise, and time required \cite{kim2019deep,cherian20212021,simpson2019large}. In such scenarios, semi-supervised learning (semi-SL) offers an efficient alternative, especially when there are only a few labeled samples but plenty of unlabeled or unaudited data. Semi-SL algorithms can leverage the vast pool of unaudited training data by extracting discriminative information from the structure of unlabeled data that complements the knowledge gained from a small number of supervisory data samples. However, semi-SL assumes that the underlying distribution of the unaudited data matches that of the few labeled samples \cite{yu2020multi,park2023opencos}.

The idea of the data distributions matching is frequently challenged in practical situations, particularly when the data is sourced from datasets with high heterogeneity. An example of such datasets' is medical images where the manifestation of patient characteristics and diseases can exhibit significant diversity.\cite{kurian2021sample}. Furthermore, representing disease continuums as discrete categories inherently introduces out-of-distribution (OOD) samples in medical imaging data. Consequently, the presence of OOD samples in the unlabelled training pool of semi-supervised learning (semi-SL) is inevitable. In a supervised learning framework, such OOD training samples are referred to as open-set samples, as they may not strictly conform to the distribution of any predefined classes of interest in a classification problem \cite{ luo2021empirical}.

In the case of deep semi-SL frameworks, the presence of OOD samples in the unlabeled training data can have a negative impact on knowledge propagation, thereby reducing the efficiency of the algorithm \cite{yu2020multi}. To address this issue, common preprocessing methods aim to filter out the outlier samples \cite{yu2020multi}. However, such methods may not be suitable for semi-SL frameworks that deal with medical images. Medical images often involve a wide range of anatomical structures with significant differences in their visible features, making it difficult to eliminate samples without losing valuable information. Additionally, certain disease conditions in medical images can present as rare morphologies, which may be missed by filtering out samples. Therefore, it is important to develop tailored solutions for open-set semi-supervision in medical imaging.

The focus of our work is to address the challenges of open-set supervised learning in digital histology images, where the availability of labeled data is limited but noise-free with respect to its supervision and unlabeled data contains open-set samples. We propose a pipeline that combines an OOD detector based on self-supervised learning with a semi-SL framework to effectively utilize the large pool of unaudited data for training. Our pipeline is designed specifically for medical images, which often contain novel samples that should not be naively discarded. The OOD detector's outlier score is used to modulate the sampling selection for the subsequent semi-SL stage to ensure that the samples conform to the distribution of the few labeled samples.

We conducted extensive experiments on two digital pathology datasets: Kather colorectal histology dataset \cite{kather2019predicting} and a dataset derived from TCGA-BRCA whole slide images (WSIs) \cite{weinstein2013cancer}. The Kather dataset was customized to contain varying proportions of open-set data in unlabeled samples. Our experiments show that our proposed method is effective in addressing open-set supervised learning challenges in digital histology images compared to other popular methods and frameworks in semi-SL algorithms.

Our method is compatible with any semi-SL framework, though we base our experiments on the popular Mixmatch semi-SL framework. We demonstrate that our method outperforms other semi-SL frameworks in terms of accuracy, demonstrating the effectiveness of our method in digital histology images. The pipeline can be used as a solution to address open-set supervised learning challenges in digital histology images where labeled data is limited and unlabeled data contains open-set samples.

\begin{figure*}[ht]
\centering   \includegraphics[scale=0.34]{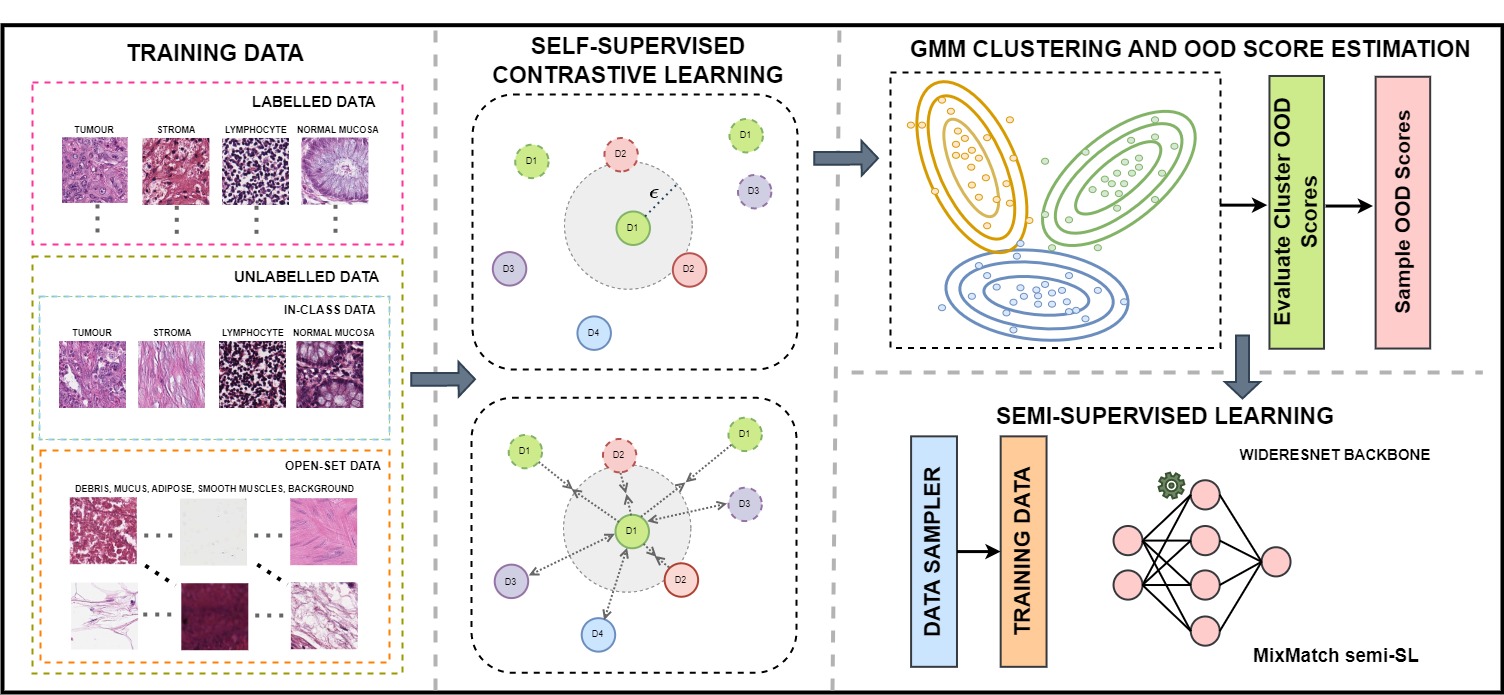}
\caption{Proposed self-supervision based open-set semi-SL framework.}
\label{fig:3_ossl_method}
\vspace{-4mm}
\end{figure*}
\section{Related Work}
Several methods have been reported in literature that developed novel algorithms that can leverage the supervisory information from labeled data and propagate it into unlabelled samples. The state-of-the-art methods in semi-SL  employ consistency regularization \cite{zhang2019consistency} to reliably extract and incorporate unsupervised information coherent with the labeled data. In consistency regularization \cite{zhang2019consistency}, the classification results for the unlabeled sample are regularized to remain unchanged with data augmentations ensuring that only robust information encoded in the structure of unlabeled data is learned. The major semi-SL algorithms that use this consistency regularization \cite{zhang2019consistency} are enlisted and concisely explained below. We also summarize a multi-task curriculum learning framework (MTL) \cite{yu2020multi} approach, which has recently been reported to successfully address the open-set issues in semi-supervised learning.

\subsubsection{MixMatch}
\label{sec:3_ossl_mm}
MixMatch \cite{berthelot2019mixmatch} is a popular semi-SL algorithm that integrated three distinct algorithmic subroutines: consistency regularization \cite{zhang2019consistency}, entropy minimization, and MixUp \cite{zhang2017mixup}, to ensure efficient training and knowledge abstraction from the labeled and unlabeled training samples. As mentioned before, consistency regularization \cite{zhang2019consistency} is employed in these algorithms to ensure that training data (including the unlabeled data) and its augmentations have similar predictions throughout the training. Entropy minimization sharpens the pseudo-labeled predictions on multiple input augmentations, encouraging the model to output more confident predictions on the unlabelled samples. Finally, by augmenting the input data with convex combinations of the training samples and their corresponding labels, the concept of MixUp \cite{zhang2017mixup} introduces a linear relationship between the training samples. The entire framework uses a cross-entropy based loss function for the fully supervised samples, whereas a mean square loss is adopted to optimize the unlabeled data.

\subsubsection{FixMatch}
FixMatch \cite{sohn2020fixmatch} is another semi-SL algorithm that works on the principles of pseudo-labeling and consistency regularization \cite{zhang2019consistency} concepts. FixMatch \cite{sohn2020fixmatch} modifies the original MixMatch \cite{berthelot2019mixmatch} framework by removing the MixUp \cite{zhang2017mixup} stage and introducing an enhanced version of consistency regularization \cite{zhang2019consistency} between weakly and strongly augmented versions of training samples. FixMatch \cite{sohn2020fixmatch} ensures superior pseudo-labeling by imposing an additional threshold value on weak augmentation so that only highly confident predictions from these samples are considered for consistency regularization.

\subsubsection{ReMixMatch}
ReMixMatch \cite{berthelot2019remixmatch} enhances the MixMatch \cite{berthelot2019mixmatch} framework by introducing the principles of distribution alignment and augmentation anchoring. The notion of distribution alignment replaces the sharpening step of the MixMatch \cite{berthelot2019mixmatch} framework in an attempt to match model aggregated class prediction to that of the marginal distribution of the given ground truth. Further, ReMixMatch \cite{berthelot2019remixmatch} introduces another principle of augmentation alignment in place of the consistency regularization  step of MixMatch \cite{berthelot2019mixmatch} to encourage each output to be close to the prediction for a weakly-augmented version of the same input. By incorporating these changes, ReMixMatch \cite{berthelot2019remixmatch} has been reported to be data efficient compared to MixMatch \cite{berthelot2019mixmatch} frameworks.

\subsubsection{Multi-task Curriculum learning (MTL) Framework Guided Semi-SL}
Unlike previous methods that focused on developing strong semi-SL models, the goal of MTL \cite{yu2020multi} guided semi-SL is to explicitly address the issues associated with the presence of open-set samples in semi-SL. The proposed framework addresses this issue by simultaneously training and optimizing OOD detection, and a MixMatch \cite{berthelot2019mixmatch}-based semi-SL in a multitask learning framework. The paper proposes a novel OOD detector based on the model's capacity to identify noisy labeled training data. The OOD detector ensures that the subsequent semi-SL framework, based on MixMatch \cite{berthelot2019mixmatch}, is only trained with the inliers samples by selecting an appropriate threshold on the OOD score and filtering out the outlier samples from the unlabelled data. The method significantly produced superior results in settings where samples from a different distribution contaminates the unlabelled data.

Along with these frameworks, a few studies have also re-examined the original MixMatch \cite{berthelot2019mixmatch} algorithm itself. One such study found that the performance degradation caused by open-set samples in unlabelled data is primarily due to the Pseudo-Labelling (PL) task of MixMatch.  Additionally, they also report that the incorporation of consistency regularization has the potential to enhance the performance of MixMatch in semi-SL, even when unlabelled training samples contain open-set data.

% In addition to frameworks such as MTL that are reported to address the degradation caused by open-set samples in unlabelled data, a few other studies have also reframed the original MixMatch \cite{berthelot2019mixmatch} framework itself. Decoupling the MixMatch  framework it has been reported that the deterioration of performance in the face of unlabelled open-set samples are primarily caused by pseudo-labelling task of the MixMatch algorithm. 

\section{Method}
In this section, we present our  robust semi-supervised learning (semi-SL) framework which involves multiple stages of deep learning-oriented data preprocessing. Our approach is designed to limit the exposure of open-set samples to a MixMatch-based semi-supervised learning framework.

The first stage involves training a deep learning (DL) architecture based on self-supervised learning to create a novel outlier detector. In the second stage, we calculate an out-of-distribution (OOD) score based on Gaussian mixture modeling. Finally, we develop a custom data sampler that modifies the exposure of unlabeled samples based on the OOD scores we calculated.

The entire methodology is summarized in Figure \ref{fig:3_ossl_method}. In the following sections, we provide a detailed explanation of each of these steps.

\subsubsection{Self-supervised learning stage}
To develop our OOD detector, we utilized self-supervised learning (SSL) which has been reported to discover the underlying structure of the data without explicit supervision \cite{ciga2022self}. To achieve this, we employed the widely used SimCLR-based self-supervised learning paradigm \cite{simclr} and tailored it to train a novel OOD detector for our purposes.

The SimCLR framework \cite{simclr} is based on contrastive learning principles where a sample and its augmented version are trained to produce similar embeddings in a high-dimensional latent space of dimension $D$ by optimizing a loss function based on cosine similarity. In order to train such networks, each mini-batch must include multiple augmented views of a training sample, and each training sample must serve as a positive anchor for all of its augmented versions during loss calculations.

In our methodology, in addition to the augmented versions of a training sample, we also encourage positive anchors to attract any samples that lie within an $\epsilon$ neighbourhood in the high-dimensional embedding space based on their cosine similarity. This ensures that samples with similar underlying structures are clustered closer together at the end of training.

This means that a sample $x_j$ is included as a positive pair with an anchor sample $x_i$ if it is an augmented version of $x_i$, or if the cosine similarity of their latent space embeddings $z_i$ and $z_j$ are similar to each other based on Equation \ref{eqn:3_ossl_sim}.

\begin{equation}
\label{eqn:3_ossl_sim}
    sim(z_i, z_j) \ge 1-\epsilon
\end{equation}

 Here $sim$ refers to the cosine similarity function as defined below in Equation \ref{eqn:3_ossl_cos}. 

\begin{equation}  \label{eqn:3_ossl_cos}
sim(z_i,z_j)=\frac{z_i^Tz_j}{\left|z_i\right|_2\left|z_j\right|_2}
\end{equation}
where $T$, represents the vector transpose operation and the notation $\left|z_i\right|_{2}$ represents the $L_2$ norm of the vector $z_i$.

 For a minibatch of size $2N$ ($N$ samples collated with their augmented versions), the loss function $\mathds{L}(z_i,z_j)$, for the latent space embeddings $(z_i,z_j)$ corresponding to a positive pair of training samples $(x_i,x_j)$  is given by the following negative log likelihood formulation over softmax as described in Equation \ref{eq:3_ossl_sim_loss}
\begin{equation} \label{eq:3_ossl_sim_loss}
    \mathds{L}(z_i,z_j) = - log\frac{exp (sim(z_i, z_j)/\tau)}{\sum_{k=1}^{2N} \mathbbm{1}_{[k\ne i]} exp (sim(z_i, z_k)/\tau)}
\end{equation}
In this equation, $\mathbbm{1}_{[k\ne i]} \in \{0,1\}$ is an indicator function evaluating to 1 iff $k \ne i$, $\tau$ denotes a temperature parameter and $k$ denotes the index of all samples that forms the negative pair to the anchor samples index $i$.

\subsubsection{GMM-based outlier score calculation}
Once we have trained the self-supervised OOD detector model, we proceed to estimate outlier scores for each unlabeled training sample by modelling latent space as a mixture of Gaussians\cite{reynolds2009gaussian}. The Gaussian mixture model is a probabilistic clustering model that assumes all data points are generated from a mixture of a finite number of Gaussian distributions with unknown parameters \cite{reynolds2009gaussian}. 

In this case, for the number of cluster components $N_{cls}$, we assume that for each cluster $c_n$, the distribution  of the $D$ dimensional latent space embedding variable $z$ is having a marginal probability distribution of the form:
\begin{equation}
    P(z)=\sum_{n=1}^{N_{cls}}\pi_n  P(z|C_i=c_n)
\end{equation}

where $i\in\{1,2,\cdots N_{cls}\}$, $P(z|C_i=c_n)$ is the likelihood of the mixture component and $\pi_n$ is 
the prior probability associated with the cluster component $c_n$.
% the mixture component and $\pi_i$ is the mixture proportion representing the probability that $Z_i$ belongs to the $c_i$ mixture component.

In GMMs, the likelihood distribution $P(z|C_i=c_n)$ is further assumed to be a multivariate $D$ dimensional normal distribution, $\mathcal{N}(z;\mu_n,\Sigma_n)$ with mean vector $\mu_n$ and covariance matrix $\Sigma_n$. 
% Finally the parameters of the GMM, are obtained via an Expectation Maximzation (EM) based iterative framework.
% distributed in a multidimensional Gaussian distribution with parameters $(\mu,\sigma)$

Finally, we use an iterative Expectation-Maximization (EM) algorithm \cite{moon1996expectation} to estimate the $\mu_n$, $\Sigma_n$ and $\pi_n$  parameters  of the GMM. The GMM-based soft clustering provides a normalized soft-cluster membership score of $\phi_k(c_j)$ in the range of $(0,1)$ for the cluster component $c_j$ as a posterior distribution from the prior $\pi_n$ and the likelihood, $\mathcal{N}(z;\mu_n,\Sigma_n)$, as defined in Equation \ref{cls_mem}
\begin{equation}
    \phi_{k}(c_{j})=\frac{\pi_{j}\mathcal{N}(z_k;\mu_{j},\Sigma_{j})}{\sum_{n=1}^{N_{cls}}\pi_n  \mathcal{N}(z_k;\mu_n,\Sigma_n)}
    \label{cls_mem}
\end{equation}
% \begin{equation}
%     \phi_{k,cj}(z_k)=\frac{\pi_{j}\mathcal{N}(z_k;\mu_{j},\Sigma_{j})}{\sum_{n=1}^{N_{cls}}\pi_n  \mathcal{N}(z_k;\mu_n,\Sigma_n)}
% \end{equation}

To estimate a cluster impurity score (CIS), $CIS(c_j)$, for the GMM cluster component $c_{j}$, we use the cumulative GMM membership of all labeled training samples to the total GMM membership of all the samples in a given cluster, as specified in Equation \ref{eqn:3_ossl_cis}. The intuition behind this approach is that a cluster is likely to have more inlier samples if they have more GMM membership affiliations from the inlier labeled training samples.

\begin{equation}
\label{eqn:3_ossl_cis}
    CIS(c_j)=-\log\frac{\sum_{l=1}^{|\mathcal{L}|}\phi_{\mathcal{L}_{l}}(c_j)}{\sum_{l=1}^{|\mathcal{L}|}\phi_{\mathcal{L}_{l}}(c_j) + \sum_{u=1}^{|\mathcal{U}|}\phi_{\mathcal{U}_{u}}(c_j)}
\end{equation}
In Equation \ref{eqn:3_ossl_cis}, $\mathcal{L}$ and $\mathcal{U}$ represents the labeled and unlabeled subsets of the training dataset. Finally, OOD score for an unlabeled sample, $x_k$, is given by the weighted sum of cluster impurity scores with sample membership scores over all the $N_{cls}$ as given in Equation \ref{eqn:3_ossl_ood}

\begin{equation}
\label{eqn:3_ossl_ood}
    OOD\text{ }Score(x_k)=\sum_{j=1}^{N_{cls}}CIS(c_j)\times \phi_k(c_j)
\end{equation}

\subsubsection{Design of OOD Score-Driven Data Sampler and MixMatch}
To prevent outlier samples from affecting the performance of the MixMatch-based semi-supervised learning framework, we employ the OOD scores computed for each unlabeled sample to guide the data selection process in each mini-batch. Specifically, we create a data sampler that prioritizes or samples more from the inlier regions of the unlabeled training data. Below are the steps to design this sampler:

\begin{steps}
\item We cluster the unlabeled samples into $N_{cls}$ using a GMM-based approach, as described before. Afterwards, we compute a cluster impurity score (CIS) using Equation \ref{eqn:3_ossl_cis}. To avoid having clusters with the same weights, we combine them to form super clusters, resulting in $N$ clusters with unique CIS values.
\item We use a weighted sampler to select one of the $N$ clusters defined in step 1. The weighting factor is the inverse of the  impurity score of each score, as this ensures that more samples will be chosen from the less impure clusters.

\item From the cluster selected in step 2, we sample an unlabeled data sample using weights that are inversely proportional to the OOD scores as defined in Equation \ref{eqn:3_ossl_ood}.
\end{steps}

We repeat steps 1 and 2 for all the samples in a mini-batch. For the final semi-supervised learning stage, we use the MixMatch framework. The MixMatch framework consists of two components in its loss function: a cross-entropy loss for the labeled training data, as given by Equation \ref{eqn:3_ossl_lmmm1}, and a consistency loss for the unlabeled data, as given by Equation \ref{eqn:3_ossl_lmmm2},

\begin{equation}
\label{eqn:3_ossl_lmmm1}
    \mathds{L}_{\mathcal{L}}=mean_{|\mathcal{L}|}(-\sum_{c=1}^C p_{sharp_c}\log (p_{model_c}(y_{li}/x_{li})))
\end{equation}

where, $p_{sharp}$ is the  sharpened prediction of labeled sample over $k$ augmentations of the input $x_{li}$ with labels $y_{li}$, $C$ denotes the total number of classes and $p_{model}$ denotes the predictions from the model.

The unlabeled loss is mean-square error defined in the Equation \ref{eqn:3_ossl_lmmm2}, 

\begin{equation}
\label{eqn:3_ossl_lmmm2}
    \mathds{L}_{\mathcal{U}}=mean_{|\mathcal{U}|}(-||q-p_{model}(x_{u})||_2)
\end{equation}

where, $q$ is the pseudo label guessed after sharpening operation by the model over $k$ augmentations, $p_{models}$ is the model's raw prediction over one unlabeled data, $x_{u}$ denotes the unlabeled input. Relating to the MixMatch \cite{berthelot2019mixmatch} theory mentioned in section \ref{sec:3_ossl_mm}, as per the Equations \ref{eqn:3_ossl_lmmm1} and \ref{eqn:3_ossl_lmmm2}, the sharpening operation incorporates entropy  minimization and the loss minimization ensures consistency regularization. 

\section{Experiments}
Our study involved a series of experiments to evaluate the effectiveness of our proposed method in comparison to existing semi-supervised learning frameworks when the unlabeled training data contained open-set samples. Additionally, we conducted experiments to determine optimal design parameters for our approach.

For our experiments, we utilized two datasets primarily. The first dataset, the Kather colorectal cancer (CRC) histopathology dataset \cite{kather2019predicting}, was publicly available. To simulate an open-set semi-SL problem, we reserved a portion of the classes in the dataset as OODs, which were solely present in the unlabeled training data. To assess the robustness of our algorithm, we varied the amount of OODs in the unlabeled training data. Our second dataset, the TCGA-BRCA cohort \cite{weinstein2013cancer}, was intended to demonstrate the effectiveness of our method in scenarios where open-set samples naturally exist in large unaudited data pools.

In all of our experiments, we utilized an OOD detector based on a SimCLR architecture using ResNet18 \cite{simclr}. We made adjustments to the ResNet18 model by eliminating the fully-connected layers after the global average pooling layer and appended a linear layer with dimensions of $128\times1$. The latent space embeddings that perform the SimCLR self-supervision \cite{simclr} were obtained from this linear layer. Additionally, this ResNet model \cite{he2016deep} had already undergone pre-training on a vast array of histology images utilizing SimCLR-based self-supervision \cite{ciga2022self}, making it ideal for fine-tuning on our dataset.

For all our experiments, we selected $\epsilon = 0.05$ based on empirical observations. We employed the MixMatch framework for semi-supervised learning, utilizing a wide-ResNet architecture \cite{zagoruyko2016wide}. Our data augmentations included color jitter, random horizontal and vertical flips, as well as elastic deformations.
\begin{figure*}[!]
\centering
    \includegraphics[scale=0.33]{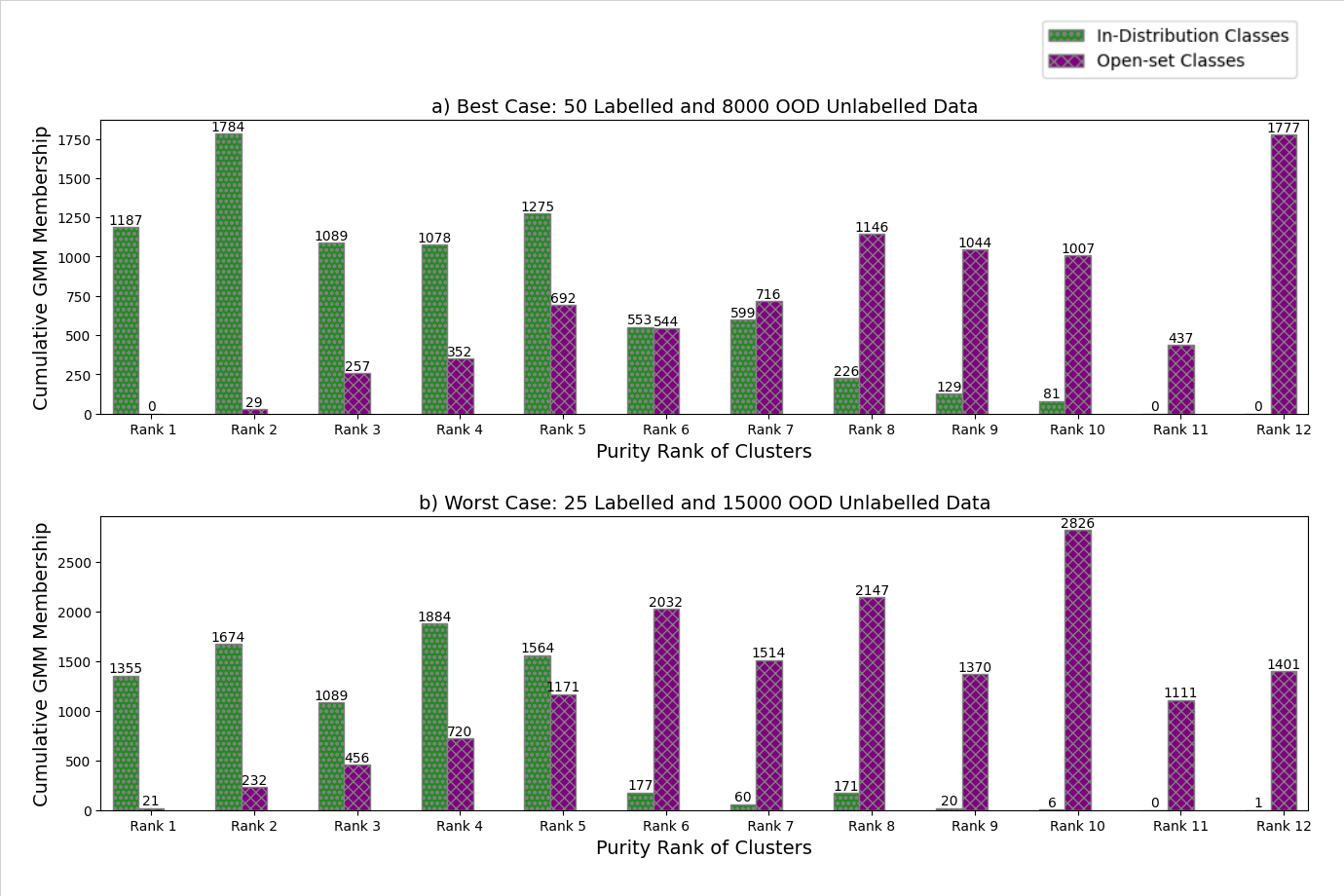}
\caption[Clusterwise distribution of inlier or OOD with 25 labeled samples.]{Cluster wise distribution of inlier or OOD samples vs purity of cluster in the experiment on Kather dataset for the best and worst case settings in our experiments.  We can see that clusters with higher CIS have more outlier samples.}
\label{fig:3_ossl_cluster25}
\vspace{-2mm}
\end{figure*}
\subsection{Experiments with colorectal cancer (CRC) Dataset}
The Kather colorectal dataset \cite{kather2019predicting} comprises 100,000 images of colorectal histology patches from H\&E-stained WSIs. These patches include nine types of tissue regions, such as adipose, background, debris, lymphocytes, mucus, normal colon mucosa, smooth muscles, cancer-associated stroma, and tumor epithelial regions. To evaluate open-set semi-supervised learning performance, we reorganized this dataset and constructed a four-class classification problem, using lymphocytes, normal colon mucosa, cancer-associated stroma, and tumor epithelial regions as the major classes. We tagged the remaining five classes as open-set samples.

To conform with the typical settings of the semi-supervised learning problem, we used only a small number of labeled samples from the four major classes for our experiments. The training data consisted mainly of unlabeled data sampled from both inlier and open-set classes. We manually varied the number of open-set samples in the unlabeled data to assess the performance drop caused by these samples. We used a labeled training data size of either 25 or 50 in each of our experiments, while keeping the number of unlabeled inlier data constant at 10,000. Additionally, we used three different sample sizes for the open-set unlabeled data: 8000, 10000, and 15000 samples.

We held out a class-balanced testing set of 4800 samples for final evaluations and also used a validation data size of 400 in our experiments.

The DNN models, including the OOD detector and MixMatch semi-SL framework, were trained using the Adam optimizer with a learning rate of 0.0003. The batch size for the OOD detector and MixMatch models was set to 32 and 128, respectively. In addition, we used 12 clusters in GMM, which was determined based on additional experiments.

%  \begin{figure}[!]
%     \centering
%     \includegraphics[scale=0.67]
%     {kather.png} 
%     \caption{Sample images from nine classes from Kather colorectal cancer histology dataset .}
%     \label{fig:intro_kather}
% % \end{figure}
% \end{figure}

\subsection{Experiments with TCGA-BRCA Whole Slide Images}

Our second set of experiments focused on classifying H\&E stained whole slide images (WSIs) based on their intrinsic subtype. We choose a Basal versus Luminal A PAM50 subtype classification for this experiment as these represent the most common histological subtype representing the worst case and best case prognosis respectively. The experiment was performed on patches extracted from WSIs and evaluated at a WSI level through patch-wise aggregation. The patches were extracted from regions of interest (tumor regions) annotated by a pathologist, which led to label corruptions due to intrinsic subtype heterogeneity in certain patients. As all patches inherited the same tumor subtype label, the label noise was more of an open-set or out-of-distribution noise, as heterogeneous tumors can exhibit a wide range of spatial cues.

The genomic criteria used to filter WSI cases for the experiment was based on a semi-supervised non-negative matrix factorization (SS-NMF) of PAM 50 genes detailed in \cite{kumar2023quantification}. Further considering quality control factors we choose a subset of 180 WSIs from the TCGA-BRCA dataset \cite{weinstein2013cancer}. This dataset was then divided into train, validation, and test sets. We selected samples with the least heterogeneity for the labeled part of the training data and for the held-out validation and test sets. As is typical in semi-SL frameworks, the number of training samples needed to be significantly smaller than the unlabeled training data. Therefore, we chose 10 WSIs (5 from Luminal A and 5 from Basal) with the most pure genomic tumor subtype signature for training, 10 WSIs for validation, and 70 WSIs for testing, all of which exhibited higher genomic tumor subtype purity compared to remaining cases. The remaining 90 WSIs, which had an admixed tumor mutational landscape, were selected as unlabeled data. In total, we extracted 50,000 patches that were approximately class-balanced across all data splits for the analysis.
\begin{table*}[t]\centering
\caption[Classification test accuracies on Kather dataset with varied levels of labelled data and OOD noise.]{Classification test accuracies on Kather dataset \cite{kather2019predicting} with varied levels of labelled data and OOD noise. Results reported here are average accuracy over three set of runs with different random seeds.}
\resizebox{2\columnwidth}{!}{
 \begin{tabular}{@{}lccccccc@{}}\toprule \midrule Labelled data & \multicolumn{3}{c}{25} & \phantom{}& \multicolumn{3}{c}{50}\\\cmidrule{2-4} \cmidrule{6-8} 
OOD Noise  & 8000    & 10000    & 15000    && 8000    & 10000    & 15000  \\ 
\midrule \midrule 

MixMatch \cite{berthelot2019mixmatch}    & 92.37$\pm$1.27 & 91.39$\pm$0.67 & 90.31$\pm$0.76 && 94.20$\pm$1.01 & 91.77$\pm$0.15    & 91.09$\pm$0.54   \\

MixMatch (w/o PL) &93.68$\pm$0.66 &91.42$\pm$0.54 &92.09$\pm$0.41 & &93.36$\pm$0.48 &92.99$\pm$0.55 &91.17$\pm$0.08 \\

FixMatch \cite{sohn2020fixmatch} & 94.13$\pm$0.26 & 90.05$\pm$0.47 & 88.46$\pm$0.61 && 94.59$\pm$0.39 & 92.50$\pm$0.20 & 91.22$\pm$0.26   \\ 

ReMixMatch \cite{berthelot2019remixmatch}  & 95.10$\pm$0.22 & 93.10$\pm$0.31 & 90.50$\pm$0.27 && 95.20$\pm$0.38 & 93.30$\pm$0.42 & 90.90$\pm$0.31  \\ 

MTL \cite{yu2020multi}  & 94.55$\pm$0.02 & 94.75$\pm$0.15 & 92.63$\pm$0.14 && 96.79$\pm$0.25 & 94.27$\pm$0.47 & 93.43$\pm$0.11  \\ 
\midrule

 Proposed Method & \textbf{97.22}$\pm$\textbf{0.29} & \textbf{96.55}$\pm$\textbf{0.20} &\textbf{ 95.58}$\pm$\textbf{0.67} && \textbf{97.82}$\pm$\textbf{0.23} &\textbf{ 97.90}$\pm$\textbf{0.06} & \textbf{97.58}$\pm$\textbf{0.46}  \\ 
 \midrule \bottomrule
\end{tabular}}
\label{tab:3_ossl_expt1}
\vspace{-3mm}
\end{table*}

\section{Results}
\subsection{Results on Kather Dataset}

\begin{figure}[!]
\centering
    \includegraphics[scale=0.2]{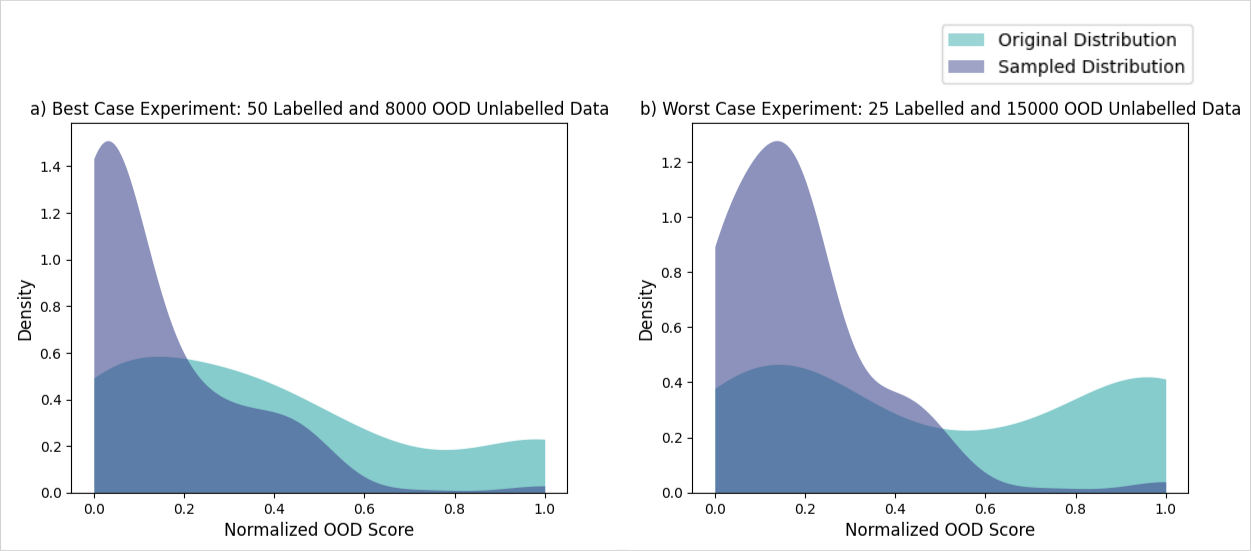}
\caption[Data sampler used in the experiments on Kather dataset.]{Data sampler used in the experiments with the Kather dataset. We can see here that this sampler chose more samples from the unlabeled training data that had less OOD scores.}
\label{fig:3_ossl_sampler}
\vspace{-4mm}
\end{figure}

After completing the training of our OOD detector, we evaluate its performance by analyzing and visualizing the distribution of GMM cluster membership for in-class and OOD samples in each of the 12 clusters. This provides us with a measure of the effectiveness of our cluster impurity score and the final OOD score. As shown in Figure \ref{fig:3_ossl_cluster25}, we observe that impure clusters with higher CIS contain more OOD samples, while pure clusters have fewer OOD samples. Figure \ref{fig:3_ossl_sampler} demonstrates how our sampling technique regulates the level of exposure of OOD samples to the following MixMatch \cite{berthelot2019mixmatch} framework.

% \begin{figure*}[tp]
% \centering
%     \includegraphics[scale=0.30]{train_50_plots.png}
% \caption[Cluster wise distribution of inlier or OOD with 50 labeled samples.]{Cluster wise distribution of inlier or OOD based on purity of cluster  in the experiment on Kather dataset with 50 labeled training samples. We can see that for clusters with higher CIS has more outlier samples.}
% \label{fig:3_ossl_cluster50}
% \end{figure*}
 
% The sampler adjusts the selection distribution of data in the unlabeled training data by favoring samples with low OOD scores.\

\begin{figure}[!]
\centering
    \includegraphics[scale=0.22]{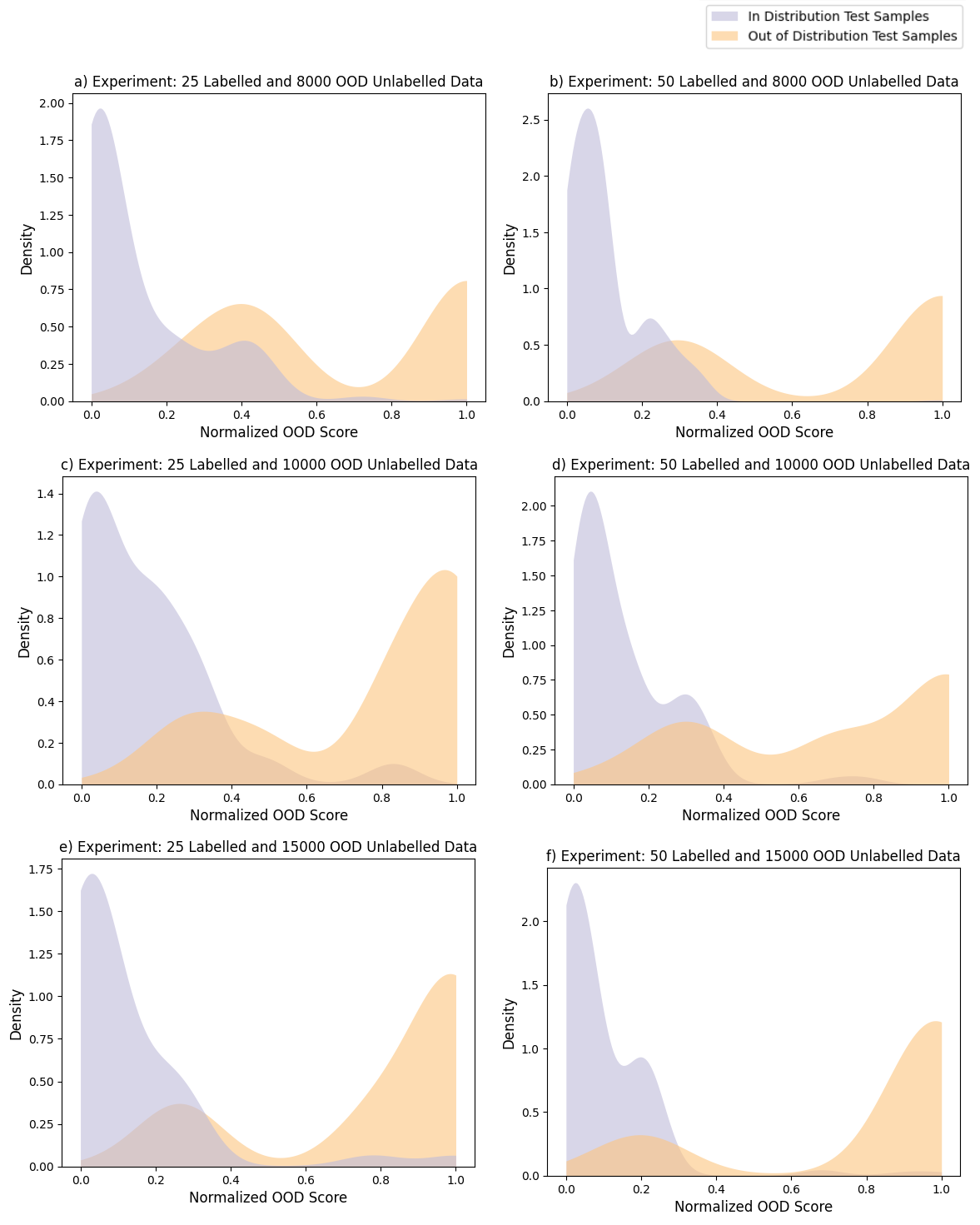}
\caption{The distribution of test samples with the OOD scores using the proposed method in Kather dataset experiments.}
\label{fig:3_ossl_ood}
\vspace{-3mm}
\end{figure}
Table \ref{tab:3_ossl_expt1} presents the main results of our experiments on the Kather dataset \cite{kather2019predicting}. It can be observed that as the number of open-set samples in the unlabeled data increases, there is a significant decrease in test accuracy. However, our proposed method demonstrates robustness to such training data corruption compared to all other baseline methods. The MTL \cite{yu2020multi} framework also performs better in these experiments. Moreover, we notice that the overall performance significantly improves when more labeled samples are included in the training data. This increase in accuracy can be attributed to the simplicity of the Kather dataset \cite{kather2019predicting} that has less intra-class variance. Therefore, more labeled samples can effectively overcome the issues associated with open-set unlabeled samples in the Kather dataset \cite{kather2019predicting}. 
Additionally, we also ran the MixMatch \cite{berthelot2019mixmatch} framework  in a scenario where there was no corruption in the unlabeled data, and we report an average accuracy over three runs to be       $98.04\%\pm0.23$ when there were 25 labelled training data and an average accuracy of $98.04\%\pm0.23$, when the labeled training data was 50.   Furthermore, the effectiveness of our proposed sampling method in calibrating the exposure of OOD samples to the subsequent MixMatch \cite{berthelot2019mixmatch} framework is illustrated in Figure \ref{fig:3_ossl_sampler}. The sampler modifies the data selection distribution in the unlabelled training data by choosing more samples from the low OOD score region.
% Additionally, Table \ref{tab:3_ossl_cleanexpt1} reports another MixMatch \cite{berthelot2019mixmatch} baseline, where there was no corruption in the unlabeled data, which forms the upper limit of accuracy achievable by our proposed method. Furthermore, the effectiveness of our proposed sampling method in calibrating the exposure of OOD samples to the subsequent MixMatch \cite{berthelot2019mixmatch} framework is illustrated in Figure \ref{fig:3_ossl_sampler}. The sampler modifies the data selection distribution in the unlabelled training data by choosing more samples from the low OOD score region.

\begin{figure*}[!]
\centering   \includegraphics[scale=0.24]{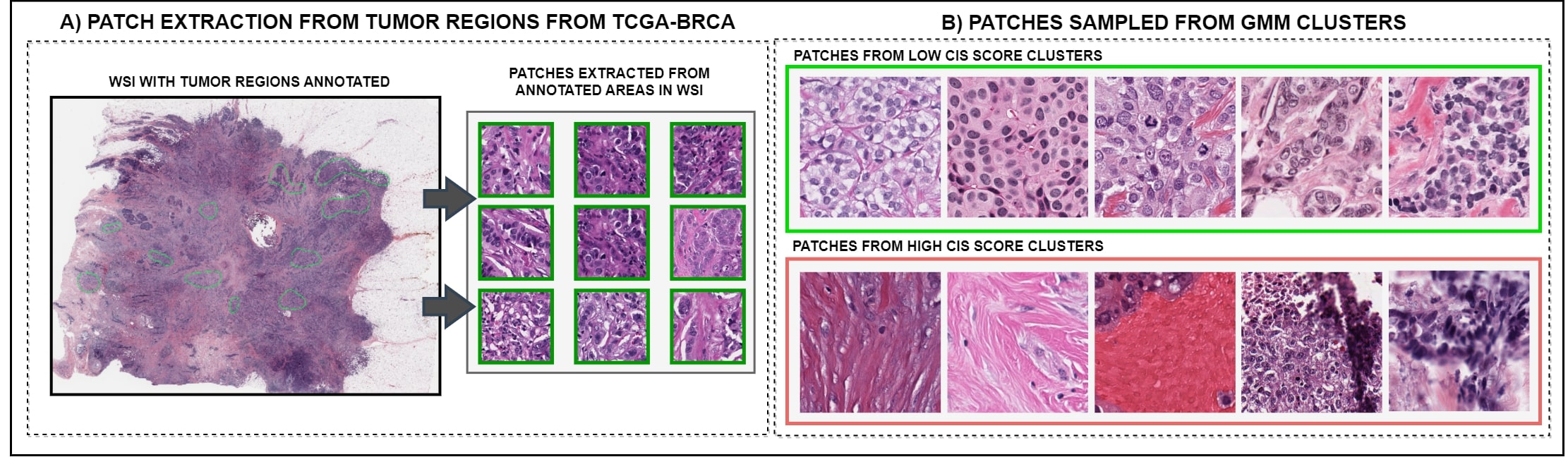}
\caption{Summary of results from experiment with RCGA-BRCA WSI dataset. Tumor region annotation and patch extraction in (A), and patch clusters with low and high impurity scores in (B). High cellularity patches are found in low CIS clusters, while high CIS clusters exhibit patches with lower cellularity or reduced label correlation due to the presence of artifacts \cite{kurian2021sample}.}
\label{expt2}
\vspace{-1mm}
\end{figure*}

% \begin{table}[ht]
% \centering
% \caption{Baseline Clean Mixmatch test accuracies on Kather dataset.}
% \begin{tabular}{@{}lcc@{}}
% \hline
% \hline
%  Labelled Data            &  25 & 50 \\
% \hline
% \hline 
% Clean Mixmatch            &   98.04$\pm$0.23 & 98.27$\pm$0.33 \\

%  \midrule \bottomrule
% \end{tabular}
% \label{tab:3_ossl_cleanexpt1}
% \end{table}

\begin{table}[ht]\centering
\caption{Comparison of OOD detection AUC scores on the test dataset of Kather dataset that had 4800 samples from inlier and OOD samples.}
\resizebox{\columnwidth}{!}{
 \begin{tabular}{@{}lccccccc@{}}\toprule \midrule Labelled data & \multicolumn{3}{c}{25} & \phantom{}& \multicolumn{3}{c}{50}\\\cmidrule{2-4} \cmidrule{6-8} 
OOD Noise  & 8000    & 10000    & 15000    && 8000    & 10000    & 15000  \\ 
\midrule \midrule 
MTL \cite{yu2020multi}  & 0.91 & 0.91 & \textbf{0.94} && 0.94 & 0.89 & 0.87  \\ 
Proposed OOD Detector & \textbf{0.92} & \textbf{0.94} & \textbf{0.94} && \textbf{0.97} & \textbf{0.93} & \textbf{0.96}  \\ 
 \midrule \bottomrule
\end{tabular}}
\label{tab:3_ossl_oodexpt1}
\end{table}

\begin{figure}[!]
\centering \includegraphics[scale=0.5]{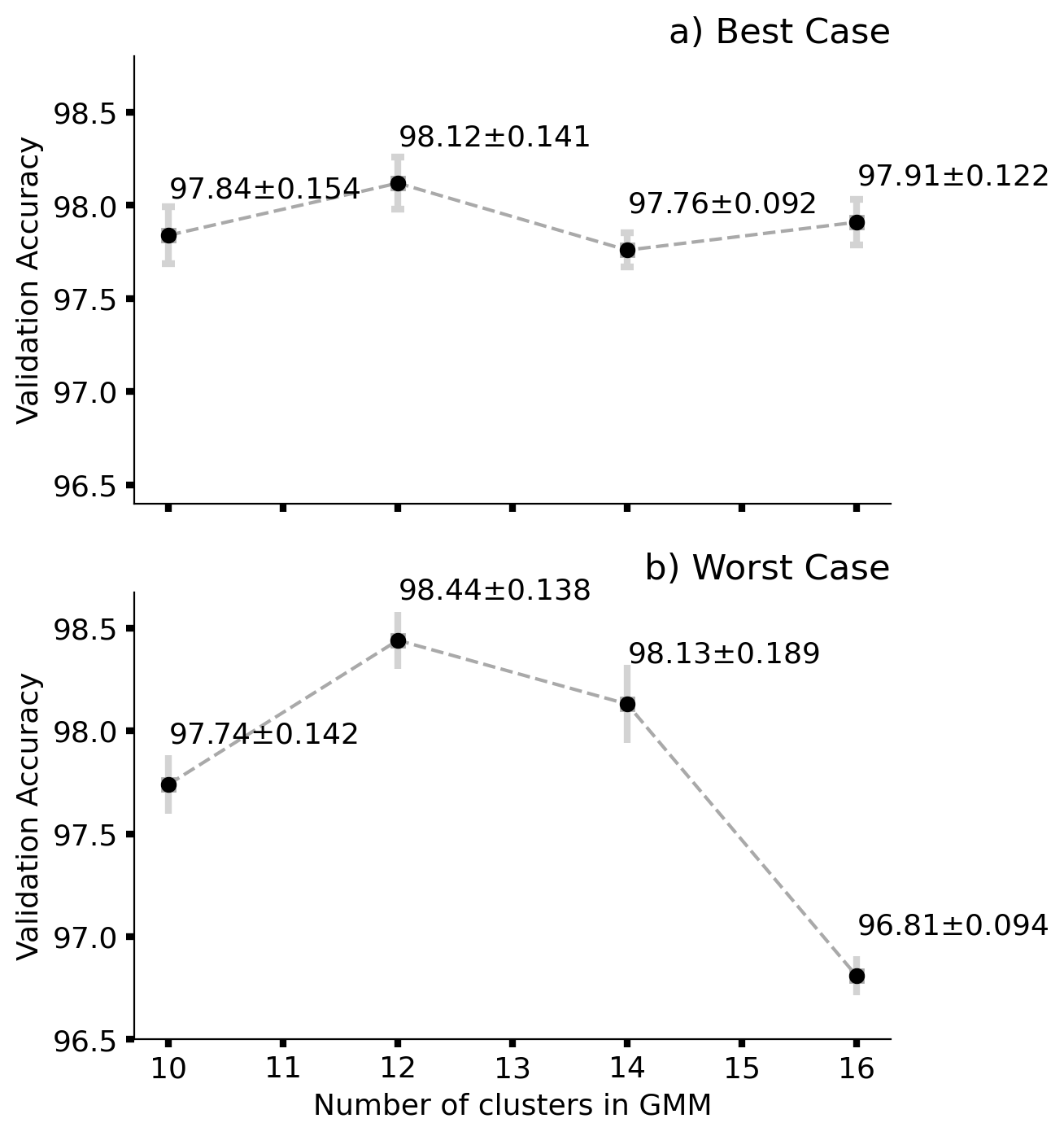}
\caption[The effect of $N_{cluster}$ on the overall performance in Kather dataset experiments for the best case scenario.]{The effect of $N_{cluster}$ on the overall performance in Kather dataset experiments for the best case scenario corresponding to 50 labeled training samples and 8000 unlabeled open-set data, and the worst case scenario in which only 25 labeled train samples and 15000 unlabeled open-set were used. Based on this study, we fixed $N_{cluster}$ to 12.}
\label{fig:3_ossl_study}
\vspace{-2mm}
\end{figure}

Additional experiments were conducted to evaluate the efficiency of our OOD detector. We plotted the distribution of test samples with the OOD scores in Figure \ref{fig:3_ossl_ood}. To further assess the OOD detector's performance, we tabulated the area under the receiver operating characteristic curve in Table \ref{tab:3_ossl_oodexpt1}. This was done on a separate test dataset consisting of 9600 samples of both OOD and inlier images, evenly sampled across all classes.

We also examined the impact of the hyperparameters, specifically $N_{clusters}$, on the overall performance, as depicted in Figure \ref{fig:3_ossl_study}. We conducted these evaluations on a dataset of 9600 instances, which consisted of 4800 OOD samples added to the inlier test data.
% \vspace{-1mm}
\begin{table}[!]\centering
\caption[Patient-level accuracy of proposed openset SSL  method for TCGA dataset]{Patient-level accuracy and patch-level shannon entropy (averaged on correctly classified patients) of proposed openset SSL  method for TCGA dataset \cite{weinstein2013cancer}}\label{tab:3_ossl_expt2}
% \scriptsize
\resizebox{\columnwidth}{!}{\begin{tabular}{lrrr}\toprule
Method &Patient Level Accuracy &Patch-Level Entropy \\\midrule
MixMatch \cite{berthelot2019mixmatch} &72.86\% &0.421 \\
MixMatch (w/o PL) &75.71\% &0.351\\
FixMatch \cite{sohn2020fixmatch} &77.14\% &0.372 \\
ReMixMatch \cite{berthelot2019remixmatch} &75.71\% &0.396 \\
MTL \cite{yu2020multi} &71.43\% &0.394 \\
\midrule
Proposed Method &\textbf{81.43\%} &0.342 \\
Proposed Method (weighted voting) &\textbf{81.43\%} &\textbf{0.305} \\
\midrule
\bottomrule
\end{tabular}}
\end{table}

Additionally, it should be noted that while our test samples were chosen to have a relatively pure genomic signature, there is still a possibility of label noise corruption in test patches. To address this issue, we implemented a modified weighted majority voting strategy that incorporated an inlier scores (1- normalized OOD scores) as weights in the patient level aggregation. However, with this modified weighting the results in Table \ref{tab:3_ossl_expt2} does not indicate any significant improvement in patient level accuracy. On the other hand, we did observe a decrease in entropy associated with patch-wise prediction when we calculated the Shannon entropy of the number of patches that were correctly classified as Luminal A and non-Luminal A. This finding indicates higher confidence in patient level predictions when the entropy is lower, which is a promising outcome for our approach. Moreover, we conducted experiments where we sampled patches from GMM clusters with low and high CIS, and our observations provide valuable insights into the effectiveness of our approach. The results of these experiments are summarized in Figure \ref{expt2}. Overall, our study contributes to the growing body of research in this field and highlights the importance of addressing label noise corruption in genomic signature analysis. We also sampled some of the patches present in the GMM clusters with low and high CIS and the observations are summarized in Figure \ref{expt2}.

% \vspace{-2mm}
\section{Conclusion}
In this paper, we proposed a novel multi-stage framework to tackle the open-set semi-supervised learning problem in histopathology. Our approach was designed to address the challenges of limited labeled data and the presence of open-set samples in the unlabeled data. Unlike other methods that use an OOD detector to discard open-set samples from the unlabeled training data, our approach limited the exposure of such samples to the semi-supervised learning framework. This was done to ensure that the training data was not artificially limited, while still mitigating the impact of potentially noisy data. 

Our proposed framework was tailored specifically for medical images, which typically have a higher degree of novelty than other types of data. We demonstrated that our framework not only preserved all the information in the data but also resulted in more robust semi-supervised learning. Our experiments showed that our approach outperformed other semi-supervised learning frameworks on two sets of histopathology images, demonstrating the effectiveness of our algorithms.

In conclusion, our multi-stage framework is a promising approach to address the open-set semi-supervised learning problem in histopathology. By limiting the exposure of open-set samples to the semi-supervised learning framework, we were able to maintain the integrity of the training data while achieving robust and effective learning. Our work has important implications for the field of medical image analysis, where limited labeled data and high novelty in unlabeled data are common challenges.

% \vspace{-1mm}
\section*{Acknowledgment}
The authors would like to thank Dr Stephanie McGregor, Assistant Professor at the University of Wisconsin School of Medicine and Public Health  for their contributions in organizing the experiment data.

\ifCLASSOPTIONcaptionsoff
  \newpage
\fi



%\bibliographystyle{IEEEtran}


% \end{thebibliography}
\bibliographystyle{IEEEbib}
\bibliography{ref}


% \end{IEEEbiography}


% % insert where needed to balance the two columns on the last page with
% %\newpage



%\vfill


\end{document}